\documentclass[runningheads]{llncs}

\usepackage[T1]{fontenc}

\usepackage{graphicx}

\usepackage{amsmath,amssymb}

\usepackage{url}

\usepackage{natbib}

\usepackage[most]{tcolorbox}

\usepackage{enumitem}

\usepackage{float}

\usepackage{multirow}
\usepackage{booktabs}  
\usepackage{tikz}      
\usepackage{array}
\usepackage{makecell}    
\usepackage{float}       
\usepackage[table]{xcolor}
\usepackage{subcaption}

\begin{document}
\title{Locating and Editing Figure-Ground Organization in Vision Transformers}
\titlerunning{Locating and Editing Figure-Ground Organization in Vision Transformers}

\author{Stefan Arnold\inst{1} \and Rene Gr\"obner\inst{1}}
\authorrunning{S. Arnold and R. Gr\"obner}

\institute{
Friedrich-Alexander-Universit\"at\\Erlangen-N\"urnberg,\\90403 N\"urnberg, Germany\\
\email{\{stefan.st.arnold, rene.edgar.groebner\}@fau.de}
}

\maketitle

\begin{abstract}

Vision Transformers must resolve figure-ground organization by choosing between completions driven by \textit{local geometric evidence} and those favored by \textit{global organizational priors}, giving rise to a characteristic perceptual ambiguity. We aim to locate where the canonical \textit{Gestalt} prior \textbf{convexity} is realized within the internal components of \texttt{BEiT}. Using a controlled perceptual conflict based on synthetic shapes of \textit{darts}, we systematically mask regions that equally admit either a \textit{concave} completion or a \textit{convex} completion. We show that \texttt{BEiT} reliably favors convex completion under this competition. Projecting internal activations into the model’s discrete visual codebook space via \textit{logit attribution} reveals that this preference is governed by identifiable functional units within transformer substructures. Specifically, we find that figure-ground organization is ambiguous through early and intermediate layers and resolves abruptly in later layers. By decomposing the direct effect of attention heads, we identify head \texttt{L0H9} acting as an early \textit{seed}, introducing a weak bias toward convexity. Downscaling this single attention head shifts the distributional mass of the perceptual conflict across a continuous decision boundary, allowing concave evidence to guide completion. 

\end{abstract}


\section{Introduction}

\emph{Vision Transformer} (ViT) \citep{dosovitskiy2021an} have redefined how visual signals are represented, modeling images as sets of interacting patches that exchange information across spatial distance. The architectural success of the ViT lies in a fundamental shift in \textit{inductive bias}. Unlike convolutional models \citep{he2016deep}, which impose a local receptive field through spatially constrained filters, transformer models are based entirely on self-attention \citep{vaswani2017attention}. Through flexible allocation of attention, ViTs maintain a global receptive field that facilitates \textit{long-range interactions} across an entire image.

Consistent with this architectural capacity, ViTs show reduced texture bias compared to convolutional models \citep{geirhos2018imagenettrained} and fostered reliance on shape information \citep{tuli2021convolutional}. Despite these architectural capabilities to integrate local cues into global shapes, it remains a subject of intense debate whether ViTs internalize fundamental principles of perceptual organization.

The laws of the organization by which coherent objects arise from the spatial configuration of spatially dispersed cues are commonly characterized through the lens of \textit{perceptual grouping} \citep{wertheimer1938laws} and \textit{figure-ground organization} \citep{wagemans2012review}. Informally, perceptual grouping explains the tendency to perceive a unified \textit{whole} rather than a fragmented \textit{mosaic}, and its process is governed by a set of well-established priors. Recent evidence suggests that model activations are indeed aligned with spatial priors such as \textit{closure} \citep{kim2021neural, zhang2025finding}, \textit{continuity} \citep{liu2022neural, biscione2023mixed}, \textit{proximity} \citep{biscione2023mixed, li2025local}, \textit{symmetry} \citep{amanatiadis2018understanding}, and \textit{similarity} \citep{amanatiadis2018understanding}. These priors capture the driving forces behind how visual elements are integrated. 

While perceptual grouping explains the integration of disjointed parts into coherent forms, figure-ground organization describes the functional segmentation of a visual scene into a salient foreground (the \textit{figure}) and a receding background (the \textit{ground}). Effectively, figure-ground organization assigns contour ownership and determines which side of a boundary forms the foreground object and which constitutes the background surface \citep{von2015figure}. This segregation is primarily driven by \textbf{convexity}. Humans exhibit a pervasive preference for perceiving convex regions as figures while relegating concave boundaries to the ground \citep{kanizsa1976convexity, peterson2008inhibitory}, and this bias has recently been observed to emerge in vision transformers \citep{li2025local}.

\begin{figure}[t]
    \centering
    \includegraphics[width=0.45\textwidth]{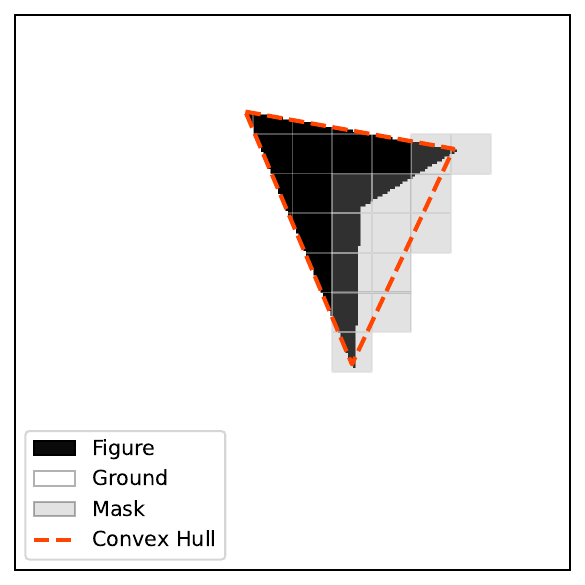}
    \caption{\textbf{A visual stimulus of perceptual conflict.} We define the conflict region (gray overlay) of interest by masking the \textit{patch}-wise set difference between the concave shape (dark region) and its convex hull (red outline), inducing competition between local evidence and global priors.}
    \label{fig:schema}
\end{figure}

\paragraph{Contribution.} We move beyond behavioral observations to provide a \emph{mechanistic basis} for figure-ground organization. We formalize this organizational paradigm as contour completion, depicted in Figure \ref{fig:schema}. Specifically, we task \texttt{BEiT} \citep{bao2022beit} with filling in a masked region within a \emph{dart}-like shape, introducing a conflict stimulus where the local concave shape competes against the global convex hull. When the model reconstructs the notch, it signals a concavity preference. Conversely, when it closes a triangle, it indicates a convexity preference. By operationalizing this preference as an internal competition, we can decouple segregation based on \textit{concave evidence} from the \textit{convexity prior}. 
\par\medskip
Our contributions are twofold:

\begin{enumerate}
    \item We decompose the residual stream of \texttt{BEiT} using \textit{logit attribution} \citep{ghandeharioun2024patchscopes} to isolate substructures responsible for figure-ground organization. We map the resolution of our figure-ground conflict stimulus to a discrete set of attention heads, demonstrating that figure-ground organization is an identifiable operation within the attention subspace.
    
    \item We then apply \textit{activation scaling} \citep{merullo2023characterizing} to intervene on figure-ground organization under masking. By modulating the functional contribution of a single influential attention head, we can reliably shift the completion tendencies, effectively flipping a convex completion into a concave interpretation. This proves that convexity prior is not a passive artifact but an active force within the model’s visual reasoning.
    
\end{enumerate}

\section{Related Work}

\subsection{Gestalt Laws in Human Perceptual Organization}

The foundation of \textit{Gestalt} psychology rests on the principle that human vision does not perceive a scene as a collection of individual cues, but rather organizes these cues into coherent wholes \citep{wertheimer1938laws}. Central to this organization are the \textit{laws of grouping}, which include aspects such as \textit{closure}, \textit{continuity}, \textit{proximity}, \textit{symmetry}, and \textit{similarity}. These laws serve as perceptual priors that allow the brain to resolve structural ambiguities. A particularly dominant prior is \textbf{convexity}, serving as a fundamental cue for \textit{figure-ground organization} \citep{kanizsa1976convexity}. In human vision, convex regions are significantly more likely to be assigned border ownership, effectively being perceived as the \textit{figure} rather than the \textit{ground} \citep{wagemans2012review, peterson2008inhibitory}.

\subsection{Emergent Gestalt Organization in Neural Networks}

A line of research has sought to determine whether the \textit{inductive biases} of neural networks align with \textit{Gestalt} principles. Early studies using convolutional models suggested a significant \textit{texture bias}, favoring local patterns over global shape \citep{baker2018deep, geirhos2018imagenettrained}. However, with the advent of the transformer architecture in vision, evidence has shifted toward a more shape-centric representation that closely parallels human perception \citep{tuli2021convolutional}. Several \textit{Gestalt} laws have been identified as emergent properties in vision models. \citet{kim2021neural} found that vision models trained on natural scenes can fill in missing contours to recognize incomplete shapes, indicating the presence of the \textit{law of closure}, although its degree remains a subject of active debate \citep{zhang2025finding}. Similarly, vison models have demonstrated an ability to recognize dashed curves by the \textit{law of continuity} \citep{liu2022neural} but have shown mixed evidence for grouping via the \textit{law of proximity} \citep{biscione2023mixed}.

More recently, vision models have been shown to integrate local cues into global percepts, exhibiting a clear \textbf{law of convexity} in line with human figure-ground tendencies \citep{li2025local}. We build on this emergent property in vision transformers. By applying techniques from the field of mechanistic interpretability to the convexity prior, we aim to target the internal computations in \texttt{BEiT} that directly govern figure–ground assignment. This provides evidence of whether transformer-based vision models implement organizational reasoning through identifiable substructures within their representation manifold.

\section{Perceptual Conflict for Mechanistic Analysis}

To investigate the competition between local concave evidence and global convexity prior, we designed a conflict stimulus based on the geometry of \textit{dart} shapes (a non-convex quadrilateral). We synthesized a sample set of 10,000 binary images containing darts. Each sample is generated with randomized position, orientations, scales, and angles to ensure the model’s response is invariant to coordinates. For each sample, we analytically determine the convex hull and define the conflict region by the geometric complement of the concave dart relative to its convex hull. Mathematically, the conflict region $M$ is defined by the set difference between the shape's convex hull $\mathcal{H}(S)$ and the shape $S$ itself: $M = \mathcal{H}(S) \setminus S$. This region represents the ambiguity where figure-ground assignment occurs.  

We operationalize this conflict stimulus by masking the patches corresponding to $M$, forcing the model to perform a \textit{zero-sum contour prediction}, where completing the triangle structure implies a convex interpretation, while preserving the dart structure is interpreted as a concave interpretation. For our masked modeling objective, we employ \texttt{BEiT} \citep{bao2022beit}. Unlike continuous autoencoder reconstruction such as \texttt{MAE}, which regresses masked pixel values, \texttt{BEiT} operates on a discrete variational autoencoder \citep{ramesh2021zero} to map masked patches to a finite codebook. This discretization allows us to treat the shape completion as a tractable classification problem.  By projecting internal activations directly onto the codebook logits, we can precisely measure the contribution of transformer substructures to the convex versus concave adjudication. The discrete output space further encourages the model to commit to a distinct hypothesis without the noise of pixel variance, rendering it a more naturalistic fit for studying the categorical nature of figure-ground organization.

\section{Model Dissection via Logit Attribution}

To isolate the model components driving the preference for convex figure in contrast to concave ground completion, we employ \textit{logit attribution} \citep{ghandeharioun2024patchscopes}, a technique originally developed to pinpoint circuits in language models \citep{wang2023interpretability}. The idea behind logit attribution is to interpret the role of a particular component in the transformer architecture in the logits space. This is grounded in the premise that the residual stream can be decomposed into the sum of contributions from every model component. Architecturally, the residual stream of a transformer acts as a communication channel. Each model component performs a \textit{read-process-write} operation, reading from the residual stream and computing an update that is linearly added back into the residual stream. Due to this additive compositionality, the input to the unembedding head (after the final layer normalization) is a linear superposition of outputs from all preceding components. This property allows us to trace the direct effect of any specific component on the logits, effectively isolating its independent vote.

We adapt this technique to \texttt{BEiT} by leveraging its \textit{discrete visual codebook} $\mathcal{V}$. Since \texttt{BEiT} predicts discrete tokens from a finite visual codebook, we can identify two distinct sets of visual tokens for our masked region: the \textit{figure set} ($\mathcal{T}_{figure}$), representing the set of visual tokens corresponding to the black pixels of the convex completion, and the \textit{ground set} ($\mathcal{T}_{ground}$), representing the set of visual tokens corresponding to the white pixels of the concave background. We then define a latent direction in the residual stream as the difference between the set's codebook vectors. For a specific component $c$, we isolate this direction by projecting its output $o_{l,h}$ through the model's codebook matrix $W_\mathcal{V}$ so that:

\begin{equation}
\text{Attribution}(c) = (\mathbf{o}_{l,h}^\top W_\mathcal{V})[\mathcal{T}_{\text{figure}}] - (\mathbf{o}_{l,h}^\top W_\mathcal{V})[\mathcal{T}_{\text{ground}}
\end{equation}


This projection yields a scalar score that reveals the figure-ground preference. A positive value indicates that the component broadcasts to the residual stream in a direction favoring convexity, while a negative value favors concavity.

\begin{figure}[htbp]
    \centering
    \begin{subfigure}{0.48\textwidth}
        \centering
        \includegraphics[width=\linewidth]{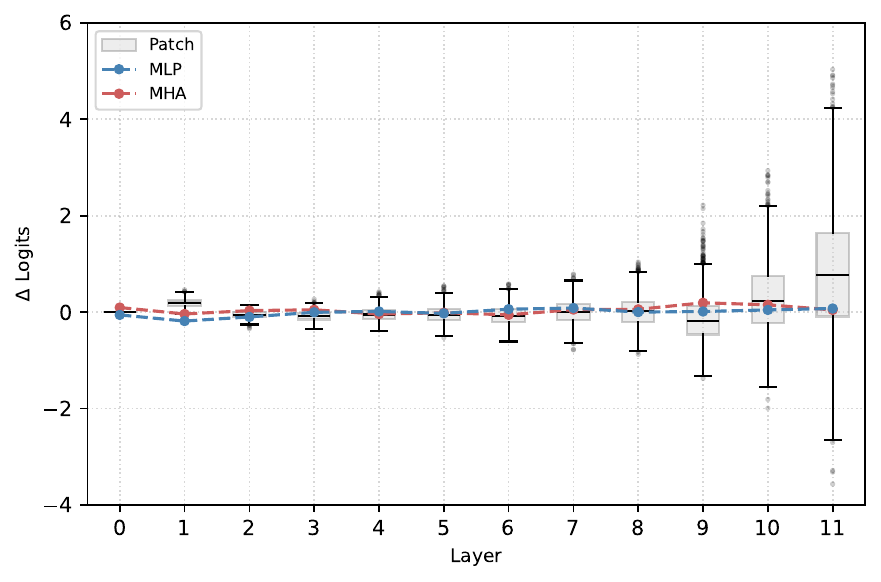}
        \caption{Logit Attribution}
        \label{fig:dla_layers}
    \end{subfigure}
    \hfill
    \begin{subfigure}{0.48\textwidth}
        \centering
        \includegraphics[width=\linewidth]{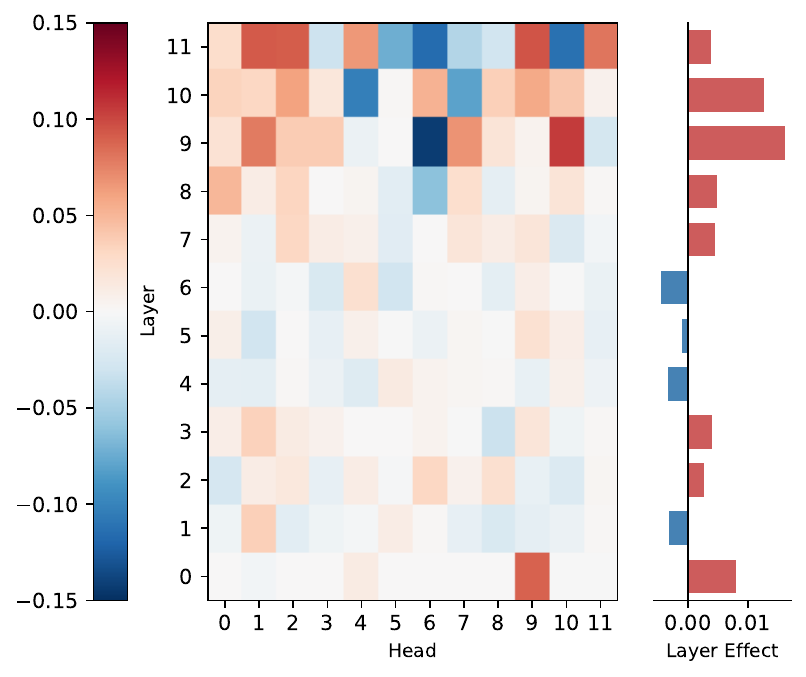}
        \caption{Attention Lens}
        \label{fig:dla_heads}
    \end{subfigure}

    \caption{\textbf{Figure--ground preference across layers and heads.} (a) Layer-wise logit attribution of the masked patch representation, showing signed residual-stream contributions favoring convex (positive) or concave (negative) completion across depth; points show medians with variability. (b) Head-wise logit attribution for attention, encoding the direction and magnitude of each head’s contribution, with marginal summaries indicating aggregate effects.}
    \label{fig:dla}
\end{figure}

\subsection{Logit Attribution and Bias Emergence}

Figure \ref{fig:dla_layers} depicts the evolution of attribution across the network depth of \texttt{BEiT}, revealing \emph{when} the model commits to a figure-ground interpretation. This analysis tests how preferences are propagated. In addition to the residual stream, we present direct effects of \textit{multi-head attention} and\textit{ multi-layer perceptron} sublayers. Since neither of these sublayers exhibit significant direct effect throughout model depth, we focus our attribution analysis on the residual stream.

Throughout the initial layers, the residual stream follows a stable trajectory. This means that the model internals maintain a state of competition. With a median attribution hovering near the zero-baseline, neither the concave evidence nor the convex prior has achieved dominance. A noticeable shift occurs only in the terminal layer. While the residual stream exhibits a clear bias toward a convex figure, the significant surge in variance (also driven by few outliers) suggests that the strength of this commitment is sensitive to the spatial configuration of the concave figure (e.g., sharpness of angles).


Since individual sublayer attributions often oscillate or partially negate one another while the final residual state shows a biased readout, we argue that this bias accumulates collectively and gradually across the entire depth of the transformer, abruptly pushing the residual stream past a decision threshold and yielding a \textit{perceptual shift}. We note that this layer-wise analysis identifies where the organizational prior becomes dominant, but falls short at explaining how directional pressure towards (against) convexity (concavity) is introduced. We address this shortcoming next by utilizing an attention lens.

\subsection{Attention Lens and Directional Pressure}

Since \citet{mehrani2023grouping} observed that attention performs perceptual grouping, we carry out a subspace attribution using the attention lens procedure established by \citet{merullo2023characterizing}. We start by extracting the corresponding vectors in the codebook matrix for the target completions. The additive update made by the attention layer is composed of the concatenated updates of each attention head after it is passed through the output weight matrix. We therefore decompose the weight matrix of the attention output into one component for each attention head and project the head activations into the space of the residual stream by multiplying them with the corresponding slice of weight matrix. By taking the dot product between each projected activation of the attention head and the target vectors in the codebook matrix, we obtain a scalar value that represents each attention head’s \textit{logit vote} toward convex or concave completion. By subtracting these two logit values, we get the direct effect to the logit difference. This difference captures the model’s internal perceptual debate.

Figure \ref{fig:dla_heads} confirms that figure-ground organization is not the outcome of unanimous agreement, but of cumulative convergence. We find that early-layer attention heads barely show any meaningful preference. Notably, \texttt{L0H9} stands out as a consistent contributor to convexity almost immediately upon input. Since the magnitude of its logit vote remains insufficient to drive commitment on its own, we assume that this attention head acts as an early \textit{seed} that introduces a subtle bias in the perceptual space toward convex interpretations. This would imply that convexity is seeded early as a weak prior rather than imposed as a hard rule. In contrast to early layer attention heads, the attention heads in later layers enter a regime of intense competition. Notably, \texttt{L9H6} acts as a concave counter-voice. Its presence indicates that geometric fidelity is not suppressed, but remains actively represented within the decision process. Despite some attention heads favoring concave completion, the ensemble effect predominantly favors convex completion. These findings indicate that convexity prevails not because concave signals disappear but because the aggregate sum of convex-supporting logit votes outweighs active opposition from concave-supporting logit votes within the attention subspaces. This confirms mechanistically that \textbf{convexity} reflects a competitive integration process in which \textit{Gestalt} priors ultimately dominate over evident boundaries.

\section{Model Steering via Activation Scaling}

We now aim to verify that our assumptions translate into a functional mechanism responsible for figure-ground organization. To provide such mechanistic proof, we transition from observation to targeted intervention. Several techniques have been proposed to edit attention activations of transformer models. \citet{merullo2023characterizing} scale the activation of an attention head by a scalar value, whereas \citet{rimsky2024steering} add a learned direction to the activation of an attention head. We adopt \textit{activation scaling} \citep{merullo2023characterizing} by applying a multiplicative scalar $\alpha$ to the activations of specific attention heads, i.e., $o_{l,h} = \alpha \cdot o_{l,h}$.

While our attribution lens highlighted several candidate attention heads, we focus our intervention on attention head \texttt{L0H9}, hypothesized to be the primary seed through which the convexity bias is injected into the residual stream. If the model’s perceptual commitment is indeed contingent upon this head's activity, then downweighting \texttt{L0H9} should effectively suppress the global prior and allow the local  evidence to dominate the final reconstruction.

Since visual tokens lack the stable mapping inherent to linguistic tokens, evaluating interventions through binary flips provides an incomplete picture of the model's confidence. Instead, we measure the probabilistic shift by calculating the \textit{Jensen-Shannon} similarity between the patch probabilities predicted by the model and our two idealized geometric targets. This formulation enables us to map the model's perceptual state onto a continuous decision manifold, capturing nuanced shifts in confidence even when the reconstruction remains unchanged.

\begin{figure}[htbp]
    \centering
    \begin{tabular}{m{0.65\textwidth} @{\hspace{0.01\textwidth}} m{0.3\textwidth}}
        
        \begin{subfigure}{\linewidth}
            \centering
            \includegraphics[width=\linewidth]{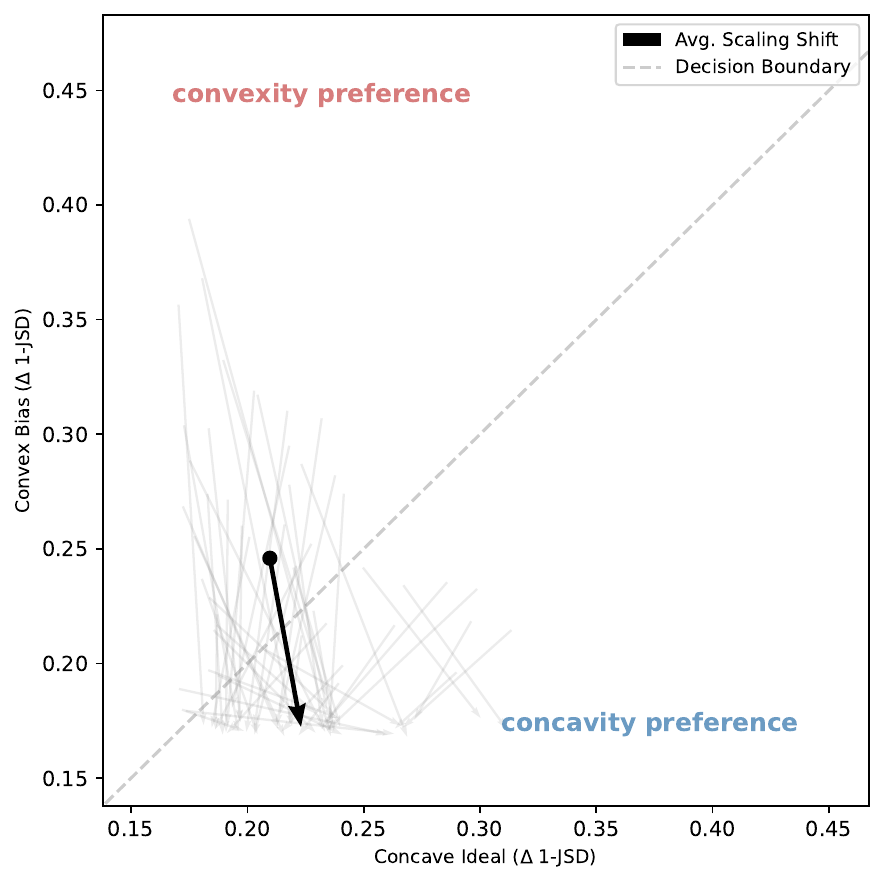}
            \caption{Activation Scaling}
            \label{fig:scaling}
        \end{subfigure}
        & 
        \begin{minipage}{\linewidth}
            \centering
            \begin{subfigure}{\linewidth}
                \centering
                \includegraphics[width=0.9\linewidth, height=2.52cm]{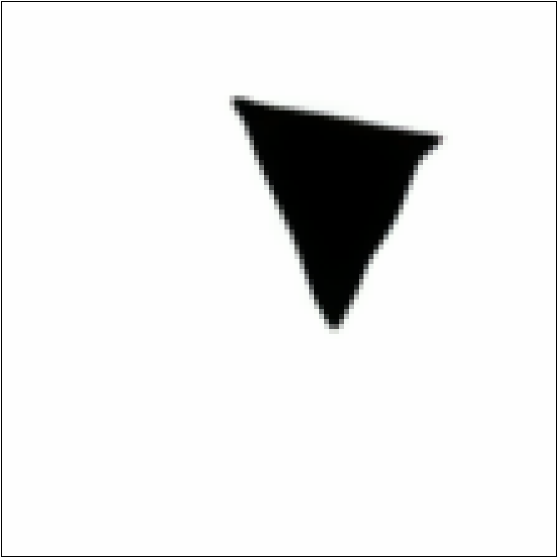}
                \caption{\textit{Convex triangle}}
                \label{fig:convex_example}
            \end{subfigure}
            \vspace{1cm} 
            \begin{subfigure}{\linewidth}
                \centering
                \includegraphics[width=0.9\linewidth, height=2.52cm]{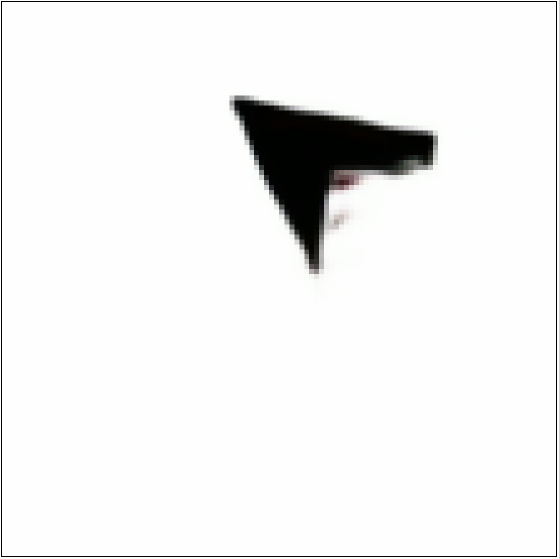}
                \caption{\textit{Concave dart}}
                \label{fig:concave_example}
            \end{subfigure}
        \end{minipage}
        
    \end{tabular}
\end{figure}

Figure \ref{fig:scaling} illustrates the results of our scaling intervention on \texttt{L0H9}, inducing a decisive shift in the model's decision-making process.\footnote{We tested multiple candidate attention heads in isolation with a multiplicate values ranging from $-2$ to $+2$ in fine-grained increments of $0.1$. This range allowed us to ablate ($\alpha=0$), amplify ($\alpha > 1$), or invert ($\alpha < 0$) the signals of an attention head.} In its nominal state of $\alpha = 1.0$, \texttt{BEiT} resides firmly in the convexity preference region of the manifold, with its probabilistic mass clustered above the diagonal decision boundary. Upon dampening the influence of \texttt{L0H9} by $\alpha = 0.3$, we observe a reliable trajectory across the decision boundary into the concavity preference zone. This shift represents a fundamental reorganization of the model’s perceptual hypothesis, where the suppression of a single early-layer head allows the concave evidence to override the convex prior. This shift is directly reflected in the visual reconstructions in Figures \ref{fig:convex_example} and \ref{fig:concave_example}. While \texttt{BEiT} by default ignores the concave notch and completes a solid triangle, the model subject to scaling intervention honors the concave shape. This transition suggests that the intervention successfully decouples the model's perceptual hypothesis from its internalized convexity prior, providing evidence that the convexity bias in vision models is not an immutable emergent \textit{Gestalt} principle, but a modifiable preference governed by mechanistic units within the attention subspace.

\section{Discussion and Conclusion}

Vision Transformers based on masked modeling objectives have been observed to internalize \textit{Gestalt} principles \citep{wertheimer1938laws} like figure-ground organization in a manner reminiscent of human perception \citep{li2025local}. We extend this line of research by proving a mechanistic understanding of how vision transformers resolve figure–ground ambiguity when local geometric evidence and global organizational priors conflict. By constructing a controlled experiment, we isolate \textbf{convexity} as a canonical prior and trace when this preference is instantiated and how it is propagated across the substructures of \texttt{BEiT} \citep{bao2022beit}. We demonstrate that convexity is an as an active organizational process mediated by identifiable attention heads, which is seeded early rather than a late-stage readout bias. By anchoring perceptual organization to concrete computational operations, this study moves \textit{Gestalt} interpretation in vision models beyond observational description toward steerable interventions.

Building on our investigation of the trajectory underlying figure-ground assignment, we postulate a \textit{seeding hypothesis}. We observe that early layers maintain a state of geometric bistability, showing no categorical preference for convex or concave completion. However, at the same time, attribution tailored to attention heads reveals a subtle yet persistent directional pressure injected by an attention head almost immediately. Since modulating this attention head fundamentally shifts the image reconstruction, we characterize this component not as a terminal decision unit, but as a primordial seed that establishes an asymmetric initial condition within the residual stream. This subtle injection of directional bias effectively skews the hypothesis space, acting as the decisive tipping point for the terminal layers, where the internal operations enter a state of intense competition between the global prior and local evidence.


\paragraph{Implications.} Beyond providing a descriptive account of model behavior, our findings have significant implications for model robustness and safety under perceptual uncertainty. Especially in domains where local concave features are diagnostically critical such as medical imaging or anomaly detection, reliability depends on the model's ability to prevent global organizational priors from catastrophically overriding local informative evidence. By demonstrating that visual organization is not an immutable architectural constraint but is instead causally governed by identifiable attention heads, we provide a framework to steer the latent decision-making process in vision transformers. This control enables the precise calibration of how models weight global priors against local evidence, ensuring more reliable object recognition in ambiguous contexts. 



\paragraph{Limitations.} We acknowledge two main limitations. \textit{First}, our mechanistic analysis focuses on a single transformer architecture, and while the identified principles of early bias seeding and late competitive commitment are supported by our results on \texttt{BEiT}, establishing generalizability requires replication across other model architectures trained with varying scales and under different regimes. \textit{Second}, our experimental setup relies on carefully designed synthetic stimuli, a deliberate choice for isolating direct effects under perceptual ambiguity. We plan to generalize our findings by extending our analysis to datasets that operationalize \textit{Gestalt} principles in more ecologically valid contexts \citep{sha2025gestaltvision}.

\bibliographystyle{plainnat}
\bibliography{custom}         

@inproceedings{he2016deep,
  title={Deep residual learning for image recognition},
  author={He, Kaiming and Zhang, Xiangyu and Ren, Shaoqing and Sun, Jian},
  booktitle={Proceedings of the IEEE conference on computer vision and pattern recognition},
  pages={770--778},
  year={2016}
}

@article{vaswani2017attention,
  title={Attention is all you need},
  author={Vaswani, Ashish and Shazeer, Noam and Parmar, Niki and Uszkoreit, Jakob and Jones, Llion and Gomez, Aidan N and Kaiser, {\L}ukasz and Polosukhin, Illia},
  journal={Advances in Neural Information Processing Systems},
  volume={30},
  year={2017}
}

@inproceedings{dosovitskiy2021an,
    title={An Image is Worth 16x16 Words: Transformers for Image Recognition at Scale},
    author={Alexey Dosovitskiy and Lucas Beyer and Alexander Kolesnikov and Dirk Weissenborn and Xiaohua Zhai and Thomas Unterthiner and Mostafa Dehghani and Matthias Minderer and Georg Heigold and Sylvain Gelly and Jakob Uszkoreit and Neil Houlsby},
    booktitle={International Conference on Learning Representations},
    year={2021},
}

@article{wertheimer1938laws,
  title={Laws of organization in perceptual forms.},
  author={Wertheimer, Max},
  year={1938},
  publisher={Kegan Paul, Trench, Trubner \& Company}
}

@article{von2015figure,
  title={Figure--ground organization and the emergence of proto-objects in the visual cortex},
  author={Von der Heydt, R{\"u}diger},
  journal={Frontiers in psychology},
  volume={6},
  pages={1695},
  year={2015},
  publisher={Frontiers Media SA}
}

@inproceedings{kanizsa1976convexity,
  title={Convexity and symmetry in figure-ground organization},
  author={Kanizsa, Gaetano and Arnheim, Rudolf and Henle, Mary and Gerbino, Walter},
  booktitle={Vision and Artifact},
  year={1976}
}

@article{peterson2008inhibitory,
  title={Inhibitory competition in figure-ground perception: Context and convexity},
  author={Peterson, Mary A and Salvagio, Elizabeth},
  journal={Journal of Vision},
  volume={8},
  number={16},
  pages={4--4},
  year={2008},
  publisher={The Association for Research in Vision and Ophthalmology}
}

@article{kim2021neural,
  title={Neural networks trained on natural scenes exhibit gestalt closure},
  author={Kim, Been and Reif, Emily and Wattenberg, Martin and Bengio, Samy and Mozer, Michael C},
  journal={Computational Brain \& Behavior},
  volume={4},
  number={3},
  pages={251--263},
  year={2021},
  publisher={Springer}
}

@article{zhang2025finding,
  title={Finding Closure: A Closer Look at the Gestalt Law of Closure in Convolutional Neural Networks},
  author={Zhang, Yuyan and Soydaner, Derya and Ko{\ss}mann, Lisa and Behrad, Fatemeh and Wagemans, Johan},
  journal={Computational Brain \& Behavior},
  pages={1--13},
  year={2025},
  publisher={Springer}
}

@inproceedings{liu2022neural,
  title={Neural recognition of dashed curves with gestalt law of continuity},
  author={Liu, Hanyuan and Li, Chengze and Liu, Xueting and Wong, Tien-Tsin},
  booktitle={Proceedings of the IEEE/CVF Conference on Computer Vision and Pattern Recognition},
  pages={1373--1382},
  year={2022}
}

@article{biscione2023mixed,
  title={Mixed evidence for gestalt grouping in deep neural networks},
  author={Biscione, Valerio and Bowers, Jeffrey S},
  journal={Computational Brain \& Behavior},
  volume={6},
  number={3},
  pages={438--456},
  year={2023},
  publisher={Springer}
}

@article{li2025local, 
  title={From Local Cues to Global Percepts: Emergent Gestalt Organization in Self-Supervised Vision Models},
  author={Li, Tianqin and Wen, Ziqi and Song, Leiran and Liu, Jun and Jing, Zhi and Lee, Tai Sing},
  journal={arXiv preprint arXiv:2506.00718},
  year={2025}
}

@inproceedings{amanatiadis2018understanding,
  title={Understanding deep convolutional networks through Gestalt theory},
  author={Amanatiadis, Angelos and Kaburlasos, Vasileios G and Kosmatopoulos, Elias B},
  booktitle={2018 IEEE international conference on imaging systems and techniques (IST)},
  pages={1--6},
  year={2018},
  organization={IEEE}
}

@InProceedings{sha2025gestaltvision,
  title = 	 {Gestalt Vision: A Dataset for Evaluating Gestalt Principles in Visual Perception},
  author =       {Sha, Jingyuan and Shindo, Hikaru and Kersting, Kristian and Dhami, Devendra Singh},
  booktitle = 	 {Proceedings of The 19th International Conference on Neurosymbolic Learning and Reasoning},
  pages = 	 {873--890},
  year = 	 {2025},
  editor = 	 {H. Gilpin, Leilani and Giunchiglia, Eleonora and Hitzler, Pascal and van Krieken, Emile},
  volume = 	 {284},
  series = 	 {Proceedings of Machine Learning Research},
  month = 	 {08--10 Sep},
  publisher =    {PMLR},
  url = 	 {https://proceedings.mlr.press/v284/sha25a.html}
}

@article{baker2018deep,
  title={Deep convolutional networks do not classify based on global object shape},
  author={Baker, Nicholas and Lu, Hongjing and Erlikhman, Gennady and Kellman, Philip J},
  journal={PLoS Computational Biology},
  year={2018},
}

@inproceedings{geirhos2018imagenettrained,
title={ImageNet-trained {CNN}s are biased towards texture; increasing shape bias improves accuracy and robustness.},
author={Robert Geirhos and Patricia Rubisch and Claudio Michaelis and Matthias Bethge and Felix A. Wichmann and Wieland Brendel},
booktitle={International Conference on Learning Representations},
year={2019},
}

@article{tuli2021convolutional,
  title={Are convolutional neural networks or transformers more like human vision?},
  author={Tuli, Shikhar and Dasgupta, Ishita and Grant, Erin and Griffiths, Thomas L},
  journal={arXiv preprint arXiv:2105.07197},
  year={2021}
}

@inproceedings{bao2022beit,
    title={{BE}iT: {BERT} Pre-Training of Image Transformers},
    author={Hangbo Bao and Li Dong and Songhao Piao and Furu Wei},
    booktitle={International Conference on Learning Representations},
    year={2022},
}

@inproceedings{ramesh2021zero,
  title={Zero-shot text-to-image generation},
  author={Ramesh, Aditya and Pavlov, Mikhail and Goh, Gabriel and Gray, Scott and Voss, Chelsea and Radford, Alec and Chen, Mark and Sutskever, Ilya},
  booktitle={International Conference on Machine Learning},
  year={2021},
}

@inproceedings{wang2023interpretability,
    title={Interpretability in the Wild: a Circuit for Indirect Object Identification in {GPT}-2 Small},
    author={Kevin Ro Wang and Alexandre Variengien and Arthur Conmy and Buck Shlegeris and Jacob Steinhardt},
    booktitle={The Eleventh International Conference on Learning Representations },
    year={2023},
}

@inproceedings{ghandeharioun2024patchscopes,
  title={Patchscopes: A Unifying Framework for Inspecting Hidden Representations of Language Models},
  author={Ghandeharioun, Asma and Caciularu, Avi and Pearce, Adam and Dixon, Lucas and Geva, Mor},
  booktitle={International Conference on Machine Learning},
  year={2024},
}

@inproceedings{merullo2023characterizing,
  title={Characterizing Mechanisms for Factual Recall in Language Models},
  author={Merullo, Jack and Yu, Qinan Jack and Pavlick, Ellie},
  booktitle={Proceedings of the 2023 Conference on Empirical Methods in Natural Language Processing},
  pages={9924--9959},
  year={2023}
}

@inproceedings{rimsky2024steering,
  title={Steering Llama 2 via Contrastive Activation Addition},
  author={Rimsky, Nina and Gabrieli, Nick and Schulz, Julian and Tong, Meg and Hubinger, Evan and Turner, Alexander},
  booktitle={Proceedings of the 62nd Annual Meeting of the Association for Computational Linguistics},
  pages={15504--15522},
  year={2024}
}

@article{wagemans2012review,
  title={A century of Gestalt psychology in visual perception: I. Perceptual grouping and figure-ground organization.},
  author={Johan Wagemans and James H. Elder and Michael Kubovy and Stephen E. Palmer and Mary A. Peterson and Manish Singh and R{\"u}diger von der Heydt},
  journal={Psychological bulletin},
  year={2012},
  volume={138 6},
  pages={
          1172-217
        },
  url={https://api.semanticscholar.org/CorpusID:10105111}
}

@misc{mehrani2023grouping,
      title={Self-attention in Vision Transformers Performs Perceptual Grouping, Not Attention}, 
      author={Paria Mehrani and John K. Tsotsos},
      year={2023},
      eprint={2303.01542},
      archivePrefix={arXiv},
      primaryClass={cs.CV},
      url={https://arxiv.org/abs/2303.01542}, 
}

\end{document}